\documentclass[10pt,twocolumn,letterpaper]{article}

\usepackage{cvpr}
\usepackage{times}
\usepackage{epsfig}
\usepackage{graphicx}
\usepackage{amsmath}
\usepackage{amssymb}

\usepackage[ruled,lined,boxed,linesnumbered]{algorithm2e}
\usepackage{multirow}
\usepackage{multicol}
\usepackage{subfigure}

\cvprfinalcopy 

\def\cvprPaperID{418} 

\begin{document}

\title{Thoracic Disease Identification and Localization with Limited Supervision}

\author{Zhe Li$^1$\thanks{This work was done when Zhe Li and Mei Han were at Google.}, Chong Wang$^3$, Mei Han$^{2*}$, Yuan Xue$^3$, Wei Wei$^3$, Li-Jia Li$^3$, Li Fei-Fei$^3$\\
$^1$Syracuse University, $^2$PingAn Technology, US Research Lab, $^3$Google Inc. \\
{\tt\small $^1$zli89@syr.edu, $^2$ex-hanmei001@pingan.com.cn,}\\
{\tt\small $^3$\{chongw, yuanxue, wewei, lijiali, feifeili\}@google.com}
}

\maketitle

\begin{abstract}
Accurate identification and localization of abnormalities from radiology images play an integral part in clinical diagnosis and treatment planning. Building a highly accurate prediction model for these tasks usually requires a large number of images manually annotated with labels and finding sites of abnormalities. In reality, however, such annotated data are expensive to acquire, especially the ones with location annotations. We need 
methods that can work well with only a small amount of location annotations. To address this challenge, we present a unified approach
that simultaneously performs disease identification and localization through the same underlying model for all images. We demonstrate that our approach 
can effectively leverage both class information as well as limited location annotation, 
 and significantly outperforms the comparative reference baseline 
in both classification and localization tasks.
\end{abstract}

\section{Introduction}
Automatic image analysis is becoming an increasingly important technique to support clinical diagnosis and treatment planning.
It is usually formulated as a classification problem where medical imaging abnormalities
are identified as different clinical conditions~\cite{shi2017multimodal,chen2015glaucoma, shin2016learning, wang2017detecting,zhang2017mdnet}.
In clinical practice, visual evidence that supports the classification result, such as spatial localization~\cite{akselrod2016region} or segmentation~\cite{zhao2016multiscale, zilly2017glaucoma} of sites of abnormalities is an indispensable part of clinical diagnosis which provides interpretation and insights. Therefore, it is of vital importance that the image analysis method is able to provide both classification results and the associated visual evidence with high accuracy.

 \begin{figure}
    \centering
    \includegraphics[width=\columnwidth]{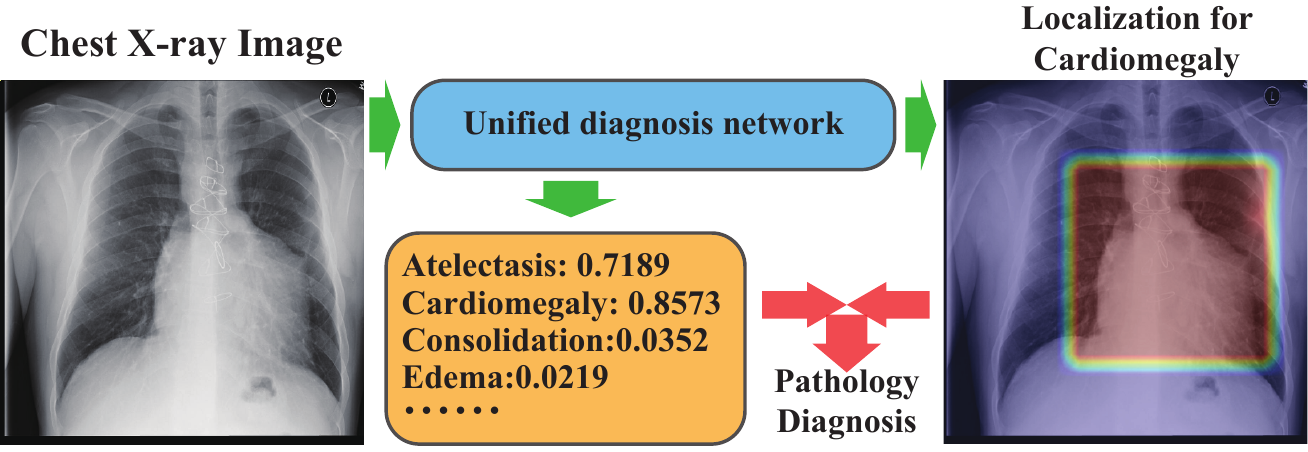} 
    \caption{Overview of our chest X-ray image analysis network for thoracic disease diagnosis. The network reads chest X-ray images and produces prediction scores and localization for the diseases}
    \label{fig:overview}
    \vspace{-1.5em}
 \end{figure}

\begin{figure}
  \centering 
  \includegraphics[width=\columnwidth]{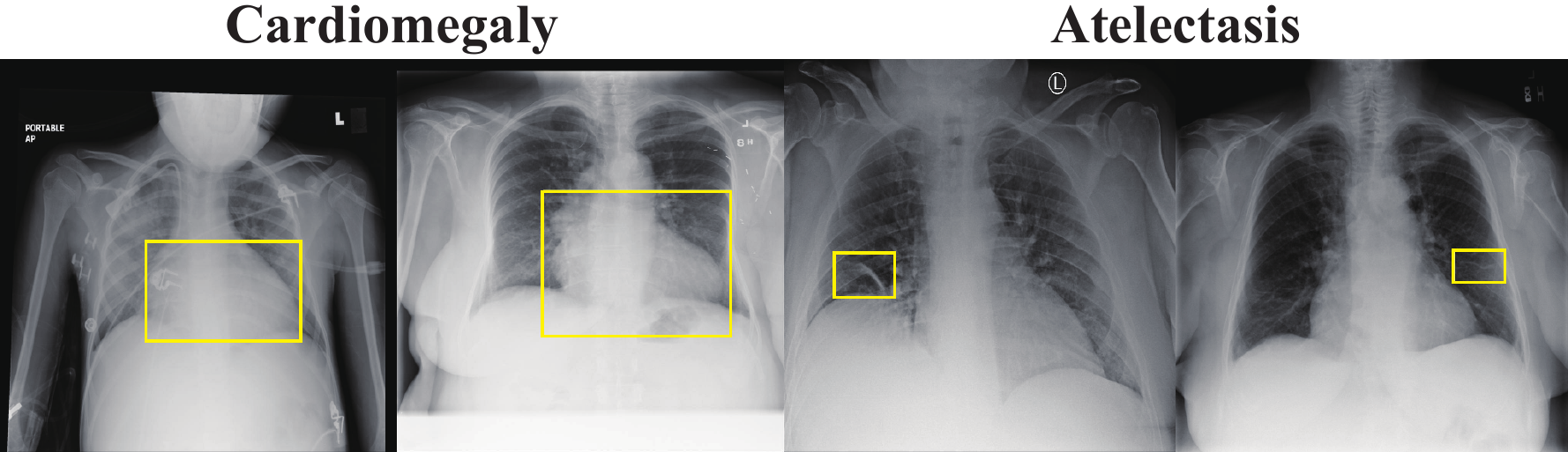}
  \caption{Examples of chest X-ray images with the disease bounding box. The disease regions are annotated in the yellow bounding boxes by radiologists. } 
  \label{fig:chestimage} 
  \vspace{-1.5em}
\end{figure}

Figure~\ref{fig:overview} is an overview of our approach. We focus on chest X-ray image analysis. Our goal is to both classify the clinical conditions and identify the abnormality locations. A chest X-ray image might contain multiple sites of abnormalities with monotonous and homogeneous image features. This often leads to the inaccurate classification of clinical conditions. It is also difficult to identify the sites of abnormalities because of their variances in the size and location.
For example, as shown in Figure \ref{fig:chestimage}, 
the presentation of ``Atelectasis'' (alveoli are deflated down) is usually limited to local regions of a lung \cite{gylys2017medical} but possible to appear anywhere on both sides of lungs; while ``Cardiomegaly'' (enlarged heart) always covers half of the chest and is always around the heart.

The lack of large-scale datasets also stalls the advancement of automatic chest X-ray diagnosis.
Wang \etal provides one of the largest publicly available chest x-ray datasets with disease labels\footnote{While abnormalities, findings, clinical conditions, and diseases have distinct meanings in the medical domain, here we simply refer them as diseases and disease labels for the focused discussion in computer vision.}
along with a small subset with region-level annotations (bounding boxes) for evaluation~\cite{wang2017chestx}\footnote{The method proposed in \cite{wang2017chestx} did not use the bounding box information for localization training.}. 
As we know, the localization annotation is much more informative than just a single disease label to improve the model performance as demonstrated in~\cite{liu2017attention}. However, getting detailed disease localization annotation can be difficult and expensive. Thus, designing models that can work well with only a small amount of localization annotation is a crucial step for the success of 
clinical applications.


In this paper, we present a unified approach that simultaneously improves disease identification and localization with only a small amount of X-ray images containing disease location information. Figure~\ref{fig:overview} demonstrates an example of the output of our model. Unlike the standard object detection task in computer vision, we do not strictly predict bounding boxes.
Instead, we produce regions that indicate the diseases, which aligns with the purpose of visualizing and interpreting the disease better. Firstly, we apply a CNN to the input image so that the model learns the information of the entire image and implicitly encodes both the class and location information for the disease~\cite{redmon2016you}. We then slice the image into a patch grid to capture the local information of the disease.
For an image with bounding box annotation, the learning task becomes a fully supervised problem since the disease label for each patch can be determined by the overlap between the patch and the bounding box. For an image with only a disease label, the task is formulated as a multiple instance learning (MIL) problem~\cite{babenko2008multiple}---at least one patch in the image belongs to that disease. If there is no disease in the image, all patches have to be disease-free. In this way, we have unified the disease identification and localization into the same underlying prediction model but with two different loss functions. 

We evaluate the model on the aforementioned chest X-ray image dataset provided in~\cite{wang2017chestx}.
Our quantitative results show that the proposed model achieves significant accuracy improvement over the published state-of-the-art on both disease identification and localization, despite the limited number of bounding box annotations of a very small subset of the data. In addition, our qualitative results reveal a strong correspondence between the radiologist's annotations and detected disease regions, which might produce further interpretation and insights of the diseases.

\section{Related Work}
\textbf{Object detection.}
Following the R-CNN work \cite{girshick2014rich}, recent progresses has focused on processing all regions with only one shared CNN \cite{he2014spatial,girshick2015fast}, and on eliminating explicit
region proposal methods by directly predicting the bounding boxes.
In \cite{ren2015faster}, Ren \etal developed a region proposal network (RPN) that regresses from anchors to regions of interest (ROIs).
However, these approaches could not be easily used for images without enough annotated bounding boxes.
To make the network process images much faster, Redmon \etal proposed a grid-based object detection network, YOLO, where an image is partitioned into $S\times S$ grid cells, each of which is responsible to predict the coordinates and confidence scores of $B$ bounding boxes \cite{redmon2016you}. The classification and bounding box prediction are formulated into one loss function to learn jointly. A step forward, Liu \etal partitioned the image into multiple grids with different sizes proposing a multi-box detector overcoming the weakness in YOLO and achieved better performance \cite{liu2016ssd}.
Similarly, these approaches are not applicable for the images without bounding boxes annotation.
Even so, we still adopt the idea of handling an image as a group of grid cells and treat each patch as a classification target.


\textbf{Medical disease diagnosis.}
Zhang \etal proposed a dual-attention model using images and optional texts to make accurate prediction \cite{zhang2017tandemnet}.
In \cite{zhang2017mdnet}, Zhang \etal proposed an image-to-text model to establish a direct mapping from medical images to diagnostic reports.
Both models were evaluated on a dataset of bladder cancer images and corresponding diagnostic reports.
Wang \etal took advantage of a large-scale chest X-ray dataset to formulate the disease diagnosis problem as multi-label classification, using class-specific image feature transformation \cite{wang2017chestx}. They also applied a thresholding method to the feature map visualization \cite{zeiler2014visualizing} for each class and derived the bounding box for each disease. 
Their qualitative results showed that the model usually generated much larger bounding box than the ground-truth.
Hwang \etal \cite{hwang2016self} proposed a self-transfer learning framework to learn localization from the globally pooled class-specific feature maps supervised by image labels.
These works have the same essence with class activation mapping \cite{zhou2016learning} which handles natural images.
The location annotation information was not directly formulated into the loss function in the none of these works. 
Feature map pooling based localization did not effectively capture the precise disease regions.

\textbf{Multiple instance learning.}
In multiple instance learning (MIL), an input is a labeled bag (e.g., an image) with many instances (e.g., image patches)~\cite{babenko2008multiple}. The label is assigned at the bag level.
Wu \etal assumed each image as a dual-instance example, including its object proposals and possible text annotations \cite{wu2015deep}.  The framework achieved convincing performance in vision tasks including classification and image annotation. In medical imaging domain, Yan \etal utilized a deep MIL framework for body part recognition \cite{yan2016multi}.
Hou \etal first trained a CNN on image patches and then an image-level decision fusion model by patch-level prediction histograms to generate the image-level labels \cite{hou2016patch}.
By ranking the patches and defining three types of losses for different schemes, Zhu \etal proposed an end-to-end deep multi-instance network to achieve mass classification for whole mammogram \cite{zhu2017deep}.
We are building an end-to-end unified model to make great use of both image level labels and bounding box annotations effectively.

\begin{figure*}
    \centering
    \vspace{-1em}
    \includegraphics[width=1.9\columnwidth]{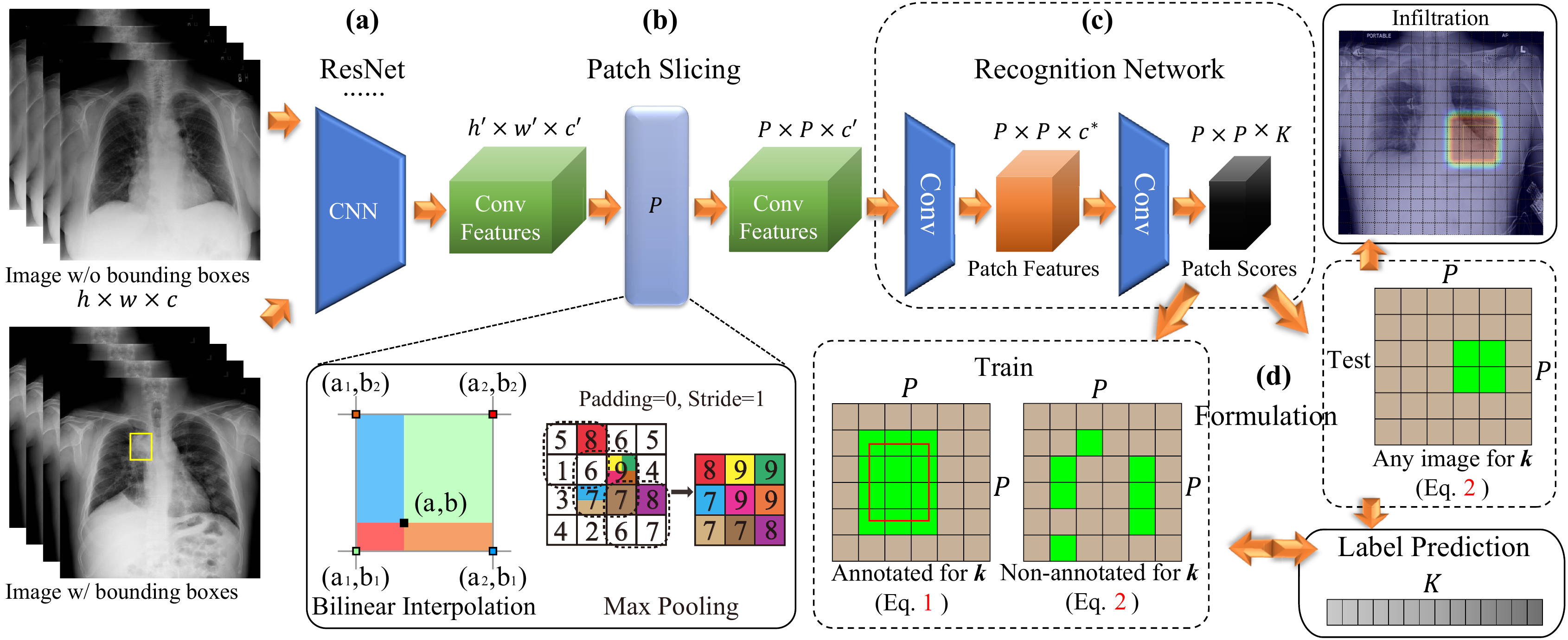} 
    \caption{Model overview. (a) The input image is firstly processed by a CNN. (b) The patch slicing layer resizes the convolutional features from the CNN using max-pooling or bilinear interpolation. (c) These regions are then passed to a fully-convolutional recognition network. (d) During training, we use multi-instance learning assumption to formulate two types of images; during testing, the model predicts both labels and class-specific localizations. The red frame represents the ground truth bounding box. The green cells represent patches with positive labels, and brown is negative. Please note during training, for unannotated images, we assume there is at least one positive patch and the green cells shown in the figure are not deterministic.}
    \label{fig:architecture}
    \vspace{-1.5em}
 \end{figure*}

\section{Model}

Given images with disease labels and limited bounding box information, we aim to design a unified model that simultaneously produces disease identification and localization. We have formulated two tasks into the same underlying prediction model so that 1) it can be jointly trained end-to-end and 2) two tasks can be mutually beneficial. The proposed architecture is summarized in Figure \ref{fig:architecture}.

\subsection{Image model}\label{subsec:imagemodel}
\textbf{Convolutional neural network.}
As shown in Figure~\ref{fig:architecture}(a), we use the residual neural network (ResNet) architecture \cite{he2016deep} given its dominant performance in ILSVRC competitions \cite{russakovsky2015imagenet}. Our framework can be easily extended to any other advanced CNN models. The recent version of pre-act-ResNet \cite{he2016identity} is used (we call it ResNet-v2 interchangeably in this paper).
After removing the final classification layer and global pooling layer, an input image with shape $h\times w\times c$ produces a feature tensor with shape $h'\times w'\times c'$ where $h$, $w$, and $c$ are the height, width, and number of channels of the input image respectively while $h'=\frac{h}{32}$, $w'=\frac{w}{32}$, $c'=2048$.
The output of this network encodes the images into a set of abstracted feature maps.

\textbf{Patch slicing.}
Our model divides the input image into $P \times P$ patch grid,
and for each patch, we predict $K$ binary class probabilities, where $K$ is the number of possible disease types.
As the CNN gives $c'$ input feature maps with size of $h' \times w'$, we down/up sample the input feature maps to $P \times P$ through a patch slicing layer shown in Figure~\ref{fig:architecture}(b).
Please note that $P$ is an adjustable hyperparameter.
In this way, a node in the same spatial location across all the feature maps corresponds to one patch of the input image.
We upsample the feature maps If their sizes are smaller than expected patch grid size.
Otherwise, we downsample them.

\textit{Upsampling.} 
We use a simple bilinear interpolation to upsample the feature maps to the desired patch grid size.
As interpolation is, in essence, a fractionally stridden convolution, it can be performed in-network for end-to-end learning and is fast and effective \cite{long2015fully}.
A deconvolution layer \cite{zeiler2014visualizing} is not necessary to cope with this simple task.

\textit{Downsampling.} 
The bilinear interpolation makes sense for downsampling only if the scaling factor is close to $1$.
We use max-pooling to down sample the feature maps. 
In general cases, the spatial size of the output volume is a function of the input width/height ($w$), the filter (receptive field) size ($f$), the stride ($s$), and the amount of zero padding used ($p$) on the border.
The output width/height ($o$) can be obtained by $\frac{w-f+2p}{s}+1$.
To simplify the architecture, we set $p=0$ and $s=1$, so that $f=w-o+1$.

\textbf{Fully convolutional recognition network.}\label{subsubsec:recognition_net}
We follow \cite{long2015fully} to use fully convolution layers as the recognition network.
Its structure is shown in Figure~\ref{fig:architecture}(c).
The $c'$ resized feature maps are firstly convolved by $3\times3$ filters into a smaller set of feature maps with $c^\ast$ channels, followed by batch normalization \cite{ioffe2015batch} and rectified linear units (ReLU) \cite{glorot2011deep}. Note that the batch normalization also regularizes the model.
We set $c^\ast=512$ to represent patch features.
The abstracted feature maps are then passed through a $1\times 1$ convolution layer to generate a set of $P \times P$ final predictions with $K$ channels. 
Each channel gives prediction scores for one class among all the patches,
and the prediction for each class is normalized by a logistic function (sigmoid function) to $[0,1]$.
The final output of our network is the $P \times P \times K$ tensor of predictions.
The image-level label prediction for each class in $K$ is calculated across $P \times P$ scores, which is described in Section~ \ref{subsec:formulation}.

\subsection{Loss function}\label{subsec:formulation}
\textbf{Multi-label classification.}
Multiple disease types can be often identified in one chest X-ray image and these disease types are not mutually exclusive. Therefore, we define a binary classifier for each class/disease type in our model. The binary classifier outputs the class probability. Note that the binary classifier is not applied to the entire image, but to all small patches. We will show how this can translate to image-level labeling below. 

\textbf{Joint formulation of localization and classification.}
Since we intend to build $K$ binary classifiers, we will exemplify just one of them, for example, class $k$. Note that $K$ binary classifiers will use the same features and only differ in their last logistic regression layers. The $i$th image $x_i$ is partitioned into a set $\mathcal{M}$ of patches equally, $x_i = [x_{i1}, x_{i2}, ..., x_{im}]$, where $m=|\mathcal{M}|=P\times P$

\textit{Images with annotated bounding boxes.} 
As shown in Figure~\ref{fig:architecture}(d), suppose an image is annotated with class $k$ and a bounding box. We denote $n$ be the number of patches covered by the bounding box, where $n<m$. Let this set be $\mathcal{N}$.
Each patch in the set $\mathcal{N}$ as positive for class $k$ and each patch outside the bounding box as negative. 
Note that if a patch is covered partially by the bounding box of class $k$, we still consider it a positive patch for class $k$. The bounding box information is not lost.
For the $j$th patch in $i$th image, let $p_{ij}^k$ be the foreground probability for class $k$. Since all patches have their labels, 
the probability of an image being positive for class $k$ is defined as,
\begin{align}\label{eqn:bbox_prod}
\small
 \textstyle  p(y_k|x_i, {\rm bbox}^k_i) = \prod_{j\in \mathcal{N}} p_{ij}^k\cdot \prod_{j \in \mathcal{M} \setminus \mathcal{N}} (1-p_{ij}^k),
\end{align}
where $y_k$ is the $k$th network output denoting whether an image is a positive example of class $k$. 
For example, for a class other than $k$, this image is treated as the negative sample without a bounding box. We define a patch as positive to class $k$ when it is overlapped with a ground-truth box, and negative otherwise.

\textit{Images without annotated bounding boxes.} If the $i$th image is labeled as class $k$ without any bounding box,
we know that there must be at least one patch classified as $k$ to make this image a positive example of class $k$.
Therefore, the probability of this image being positive for class $k$ is defined as the image-level score
\footnote{Later on, we notice an similar definition \cite{liao2017evaluate} for this multi-instance problem. We argue that our formulation is in a different context of solving classification and localization in a unified way for images with limited bounding box annotation. 
Yet, this related work can be viewed as a successful validation of our multi-instance learning based formulation.},
\begin{align}\label{eqn:nobbox_prod}
\small
 \textstyle   p(y_k|x_i) = 1-\prod_{j\in \mathcal{M}}(1-p_{ij}^k).
\end{align}
At test time, we calculate $p(y_k|x_i)$ by Eq.~\ref{eqn:nobbox_prod} as the prediction probability for class $k$.

\textit{Combined loss function.} Note that $p(y_k|x_i, {\rm bbox}^k_i)$ and $p(y_k|x_i)$ are the image-level probabilities. The loss function for class $k$ can be expressed as minimizing the negative log likelihood of all observations as follows,
\begin{small}
\begin{align}
    \mathcal{L}_k =  & \textstyle - \lambda_{{\rm bbox}}\sum_i \eta_i p(y^\ast_k|x_i, {\rm bbox}^k_i)\log(p(y_k|x_i,{\rm bbox}^k_i)) \nonumber \\ 
    &- \textstyle  \lambda_{{\rm bbox}}\sum_i \eta_i (1 - p(y^\ast_k|x_i, {\rm bbox}^k_i))\log(1-p(y_k|x_i,{\rm bbox}^k_i)) \nonumber\\ 
    & \textstyle - \sum_i (1-\eta_i) p(y^\ast_k|x_i)\log(p(y_k|x_i)) \nonumber \\
    & \textstyle - \sum_i (1-\eta_i)(1 - p(y^\ast_k|x_i))\log(1-p(y_k|x_i)), \label{eqn:loss}
\end{align}
\end{small}where $i$ is the index of a data sample, $\eta_i$ is $1$ when the $i$th sample is annotated with bounding boxes, otherwise $0$. $\lambda_{{\rm bbox}}$ is the factor balancing the contributions from annotated and unannotated samples.
$p(y^\ast_k|x_i) \in \{0,1\}$ and $p(y^\ast_k|x_i, {\rm bbox}^k_i)\in \{0,1\}$ are the observed probabilities for class $k$. Obviously, $p(y^\ast_k|x_i, {\rm bbox}^k_i)\equiv1$, thus equation \ref{eqn:loss} can be re-written as follows,
\begin{small}
\begin{align}
\mathcal{L}_k = &-\textstyle \lambda_{{\rm bbox}}\sum_i \eta_i \log(p(y_k|x_i,{\rm bbox}^k_i)) \nonumber\\ 
    & - \textstyle \sum_i (1-\eta_i) p(y^\ast_k|x_i)\log(p(y_k|x_i)) \nonumber \\
    & - \textstyle\sum_i (1-\eta_i)(1 - p(y^\ast_k|x_i))\log(1-p(y_k|x_i)). \label{eqn:loss2}
\end{align}
\end{small}
In this way, the training is strongly supervised (per patch) by the given bounding box; it is also supervised by the image-level labels if the bounding boxes are not available.

To enable end-to-end training across all classes, we sum up the class-wise loss to define the total loss as,
\begin{align*}
\small
 \textstyle   \mathcal{L} = \sum_k\mathcal{L}_k.
\end{align*}


\subsection{Localization generation}
The full model predicts a probability score for each patch in the input image. 
We define a score threshold $T_s$ to distinguish the activated patches against the non-activated ones.
If the probability score $p_{ij}^k$ is larger than $T_s$, we consider the $j$th patch in the $i$th image belongs to the localization for class $k$.
We set $T_s=0.5$ in this work.
Please note that we do not predict strict bounding boxes for the regions of disease---the combined patches representing the localization information can be a non-rectangular shape. 

\subsection{Training}\label{subsec:training}
We use ResNet-v2-50 as the image model and select the patch slicing size from $\{12,16,20\}$.
The model is pre-trained on the ImageNet 1000-class dataset \cite{deng2009imagenet} with Inception \cite{szegedy2015going} preprocessing method where the image is normalized to $[-1,1]$ and resized to $299\times299$.
We initialize the CNN with the weights from the pre-trained model, which helps the model converge faster than training from scratch.
During training, we also fine-tune the image model, as we believe the feature distribution of medical images differs from that of natural images.
We set the batch size as $5$ to load the entire model to the GPU, train the model with $500$k iterations of minibatch, and decay the learning rate by $0.1$ from $0.001$ every $10$ epochs of training data.
We add L2 regularization to the loss function to prevent overfitting.
We optimize the model by Adam \cite{kingma2014adam} method with asynchronous training on 5 Nvidia P100 GPUs. The model is implemented in TensorFlow \cite{tensorflow2015-whitepaper}.

\begin{table*}	
    \centering
    \vspace{-1.1em}
	\resizebox{1.7\columnwidth}{!}{
		\begin{tabular}{cccccccc}
		\hline\hline
		\multicolumn{1}{c|}{Disease}       & Atelectasis               & Cardiomegaly              & Consolidation             & Edema                     & Effusion                   & Emphysema                 & Fibrosis                   \\ \hline
\multicolumn{1}{c|}{baseline} & 0.70                      & 0.81                      & 0.70                      & 0.81                      & 0.76                       & 0.83                      & \textbf{0.79}                       \\ \hline
\multicolumn{1}{c|}{ours}          & $\textbf{0.80} \pm 0.00 $ & $\textbf{0.87} \pm 0.01 $ & $ \textbf{0.80}\pm 0.01 $ & $ \textbf{0.88}\pm 0.01 $ & $ \textbf{0.87}\pm 0.00 $  & $ \textbf{0.91}\pm 0.01 $ & $ 0.78\pm 0.02 $  \\ \hline
\multicolumn{1}{c|}{Disease}       & Hernia                    & Infiltration              & Mass                      & Nodule                    & Pleural Thickening         & Pneumonia                 & Pneumothorax               \\ \hline
\multicolumn{1}{c|}{baseline} & \textbf{0.87}                            & 0.66                      & 0.69                      & 0.67                      & 0.68                       & 0.66                      & 0.80                       \\ \hline
\multicolumn{1}{c|}{ours}          & $ 0.77 \pm 0.03 $         & $ \textbf{0.70}\pm 0.01 $ & $ \textbf{0.83}\pm 0.01 $ & $ \textbf{0.75}\pm 0.01 $ & $ \textbf{0.79} \pm 0.01 $ & $ \textbf{0.67}\pm 0.01 $ & $ \textbf{0.87} \pm 0.01 $ \\ \hline
		\end{tabular}
	}
\caption{AUC scores comparison with the reference baseline model. Results are rounded to two decimal digits for table readability. Bold values denote better results. The results for the reference baseline are obtained from the latest update of \cite{wang2017chestx}.}
\label{tbl:classification_comparison}
\vspace{-1.5em}
\end{table*}

\textbf{Smoothing the image-level scores.}
In Eq.~\ref{eqn:bbox_prod} and \ref{eqn:nobbox_prod}, the notation $\prod$ denotes the product of a sequence of probability terms ($[0,1]$), which often leads to the a product value of $0$ due to the computational underflow if $m=|\mathcal{M}|$ is large. The log loss in Eq.~\ref{eqn:loss} mitigates this for Eq.~\ref{eqn:bbox_prod}, but does not help Eq.~\ref{eqn:nobbox_prod}, since the log function can not directly affect its product term. 
To mitigate this effect, we normalize the patch scores $p_{ij}^k$ and $1 - p_{ij}^k$ from $[0,1]$ to $[0.98,1]$ to make sure the image-level scores $p(y_k|x_i, {\rm bbox}^k_i)$ and $p(y_k|x_i)$ smoothly varies within the range of $[0,1]$.
Since we are thresholding the image-level scores in the experiments, we found this normalization works quite well. See supplementary material for a more detailed discussion on this.

\textbf{More weights on images with bounding boxes.}
In Eq.~\ref{eqn:loss2}, the parameter $\lambda_{\rm bbox}$ weighs the contribution from the images with annotated bounding boxes. Since the amount of such images is limited, and if we treat them equally with the images without bounding boxes, it often leads to worse performance. We thus increase the weight for images with bounding boxes 
to $\lambda_{{\rm bbox}}=5$ by cross validation. 

\section{Experiments}
\textbf{Dataset and preprocessing.} 
NIH Chest X-ray dataset \cite{wang2017chestx} consists of $112,120$ frontal-view X-ray images with 14 disease labels (each image can have multi-labels). These labels are obtained by analyzing the associated radiology reports. 
The disease labels are expected to have accuracy of above $90\%$~\cite{wang2017chestx}. We take the provided labels as ground-truth for training and evaluation in this work.
Meanwhile, the dataset also contains $984$ labelled bounding boxes for $880$ images by board-certified radiologists. Note that the provided bounding boxes correspond to only 8 types of disease instances. We separate the images with provided bounding boxes from the entire dataset. Hence we have two sets of images called ``annotated'' ($880$ images) and ``unannotated'' ($111,240$ images).

We resize the original 3-channel images from $1024\times1024$ to $512\times512$ pixels for fast processing. The pixel values in each channel are normalized to $[-1,1]$. We do not apply any data augmentation techniques.

\begin{figure*}
  \centering 
  \vspace{-1em}
  \includegraphics[width=1.95\columnwidth]{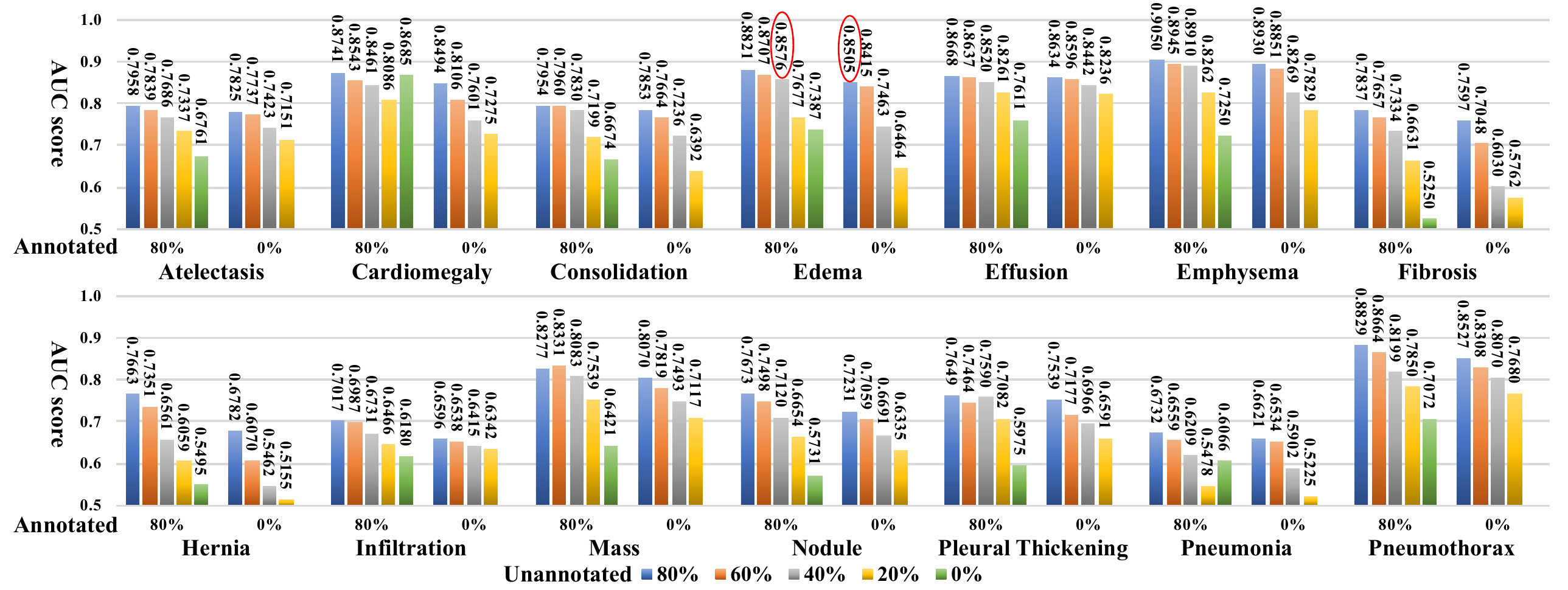}
  \caption{AUC scores for models trained using different data combinations. Training set: annotated samples,\{left: $80\%$ ($704$ images), right: $0\%$ (baseline, $0$ images)\} for each disease type; unannotated samples, \{80\% ($88,892$), 60\% ($66,744$), 40\% ($44,496$), 20\% ($22,248$), 0\%($0$)\} from left to right for each disease type. The evaluation set is $20\%$ annotated and unannotated samples which are not included in the training set. No result for 0\% annotated and 0\% unannotated images. Using $80\%$ annotated images and certain amount of unannotated images improves the AUC score compared to using the same amount of unannotated images (same colored bars in two groups for the same disease), as the joint model benefits from the strong supervision of the tiny set of bounding box annotations.} 
  \label{fig:classification} 
  \vspace{-1.5em}
\end{figure*}
\subsection{Disease identification}\label{subsec:classification}
We conduct a 5-fold cross-validation.
For each fold, we have done two experiments. 
In the first one, we train the model using $70\%$ of bounding-box annotated and $70\%$ unannotated images to compare the results with the reference model~\cite{wang2017chestx} (Table~\ref{tbl:classification_comparison}). To our knowledge, the reference model has the published state-of-the-art performance of disease identification on this dataset.
In the second experiment, we explore two data ratio factors of annotated and unannotated images to demonstrate the effectiveness of the supervision provided by the bounding boxes (Figure~\ref{fig:classification}).
We decrease the amount of images without bounding boxes from $80\%$ to $0\%$ by a step of $20\%$. And then for each of those settings, we train our model by adding $80\%$ or none of bounding-box annotated images. 
For both experiments, the model is always evaluated on the fixed $20\%$ annotated and unannotated images for this fold.

\textbf{Evaluation metrics.} We use AUC scores (the area under the ROC\footnote{Here ROC is the Receiver Operating Characteristic, which measures the true positive rate (TPR) against the false positive rate (FPR) at various threshold settings ($200$ thresholds in this work).} curve) to measure the performance of our model~\cite{fawcett2006introduction}. A higher AUC score implies a better classifier.


\textbf{Comparison with the reference model.} 
Table \ref{tbl:classification_comparison} gives the AUC scores for all the classes.
We compare our results with the reference baseline fairly:
we, as the reference, use ImageNet pre-trained ResNet-50
\footnote{Using ResNet-v2~\cite{he2016identity} shows marginal performance difference for our network compared to ResNet-v1~\cite{he2016deep} used in the reference baseline.}
, after which a convolution layer follows; both works use $70\%$ images for training and $20\%$ for testing, and we also conduct a 5-fold cross-validation to show the robustness of our model.

Compared to the reference model, our proposed model achieves better AUC scores for most diseases. The overall improvement is remarkable and the standard errors are small.
The large objects, such as ``Cardiomegaly'', ``Emphysema'', and ``Pneumothorax'', are as well recognized as the reference model.
Nevertheless, for small objects like ``Mass'' and ``Nodule'', the performance is significantly improved. Because our model slices the image into small patches and uses bounding boxes to supervise the training process, the patch containing small object stands out of all the patches to represent the complete image.
For ``Hernia'', there are only $227$ (0.2\%) samples in the dataset. These samples are not annotated with bounding boxes. Thus, the standard error is relatively larger than other diseases.

\textbf{Bounding box supervision improves classification performances.}
We consider using $0\%$ annotated images as our own baseline (right groups in Figure~\ref{fig:classification}). 
We use $80\%$ annotated images (left groups in Figure~\ref{fig:classification}) to compare with the our own baseline.
We plot the mean performance for the cross-validation in Figure~\ref{fig:classification}, the standard errors are not plotted but similar to the numbers reported in Table~\ref{tbl:classification_comparison}.  
The number of $80\%$ annotated images is just $704$, which is quite small compared to the number of $20\%$ unannotated images ($22,248$).
We observe in Figure \ref{fig:classification} that for almost all the disease types, using $80\%$ annotated images to train the model improves the prediction performance (by comparing the bars with the same color in two groups for the same disease). 
For some disease types, the absolute improvement is significant ($>5\%$). 
We believe that this is because all the disease classifiers share the same underlying image model; a better-trained image model using eight disease annotations can improve all 14 classifiers' performance. Specifically, some diseases, annotated and unannotated, share similar visual features. For example, ``Consolidation'' and ``Effusion'' both appear as fluid accumulation in the lungs, but only ``Effusion'' is annotated. The feature sharing enables supervision for ``Effusion'' to improve ``Consolidation'' performance as well.


\textbf{Bounding box supervision reduces the demand of the training images.}
Importantly, it requires less unannotated images to achieve the similar AUC scores by using a small set of annotated images for training.
As denoted with red circles in Figure~\ref{fig:classification}, taking ``Edema'' as an example, using $40\%$ ($44,496$) unannotated images with $80\%$ ($704$) annotated images ($45,200$ in total) outperforms the performance of using only $80\%$ ($88,892$) unannotated images.

\textbf{Discussion.}
Generally, decreasing the amount of unannotated images (from left to right in each bar group) will degrade AUC scores accordingly in both groups of $0\%$ and $80\%$ annotated images.
Yet as we decrease the amount of unannotated images, using annotated images for training gives smaller AUC degradation or even improvement.
For example, we compare the ``Cardiomegaly'' AUC degradation for two pairs of experiments: \{annotated:80\%, unannotated:80\% and 20\%\} and \{annotated:0\%, unannotated:80\% and 20\%\}. The AUC degradation for the first group is just $0.07$ while that for the second group is $0.12$ (accuracy degradation from blue to yellow bar).

When the amount of unannotated images is reduced to $0\%$, the performance is significantly degraded. 
Because under this circumstance, the training set only contains positive samples for eight disease types and lacks the positive samples of the other six.
Interestingly, ``Cardiomegaly'' achieves the second best score (AUC = $0.8685$, the second green bar in Figure~\ref{fig:classification}) when only annotated images are trained. The possible reason is that the location of cardiomegaly is always fixed to the heart covering a large area of the image and the feature distributions for enlarged hearts are similar to normal ones.
Without unannotated samples, the model easily distinguishes the enlarged hearts from normal ones given supervision from bounding boxes. 
When the model sees hearts without annotation, the enlarged ones are disguised and fail to be recognized. As more unannotated samples are trained, the enlarged hearts are recognized again by image-level supervision (AUC from $0.8086$ to $0.8741$).

\begin{figure*}[t]
  \centering 
  \vspace{-1em}
  \includegraphics[width=1.9\columnwidth]{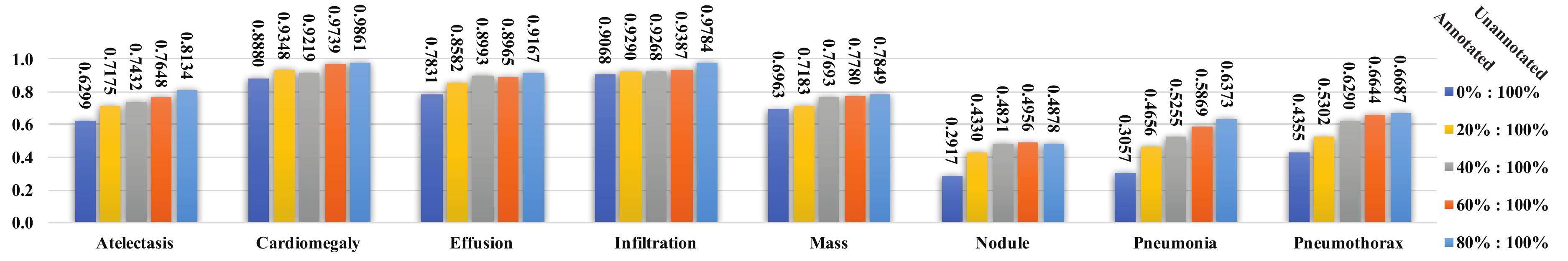}
  \caption{Disease localization accuracy using IoR where T(IoR)=0.1. Training set:  annotated samples, \{$0\%$ ($0$), $20\%$ ($176$), $40\%$ ($352$), $60\%$ ($528$), $80\%$ ($704$)\} from left to right for each disease type; unannotated samples, $100\%$ ($111,240$ images). The evaluation set is $20\%$ annotated samples which are not included in the training set. For each disease, the accuracy is increased from left to right, as we increase the amount of annotated samples, because more annotated samples bring more bounding box supervision to the joint model.} 
  \label{fig:localization_ior_1} 
  \vspace{-0.5em}
\end{figure*}
\begin{figure*}
  \centering 
  \includegraphics[width=1.9\columnwidth]{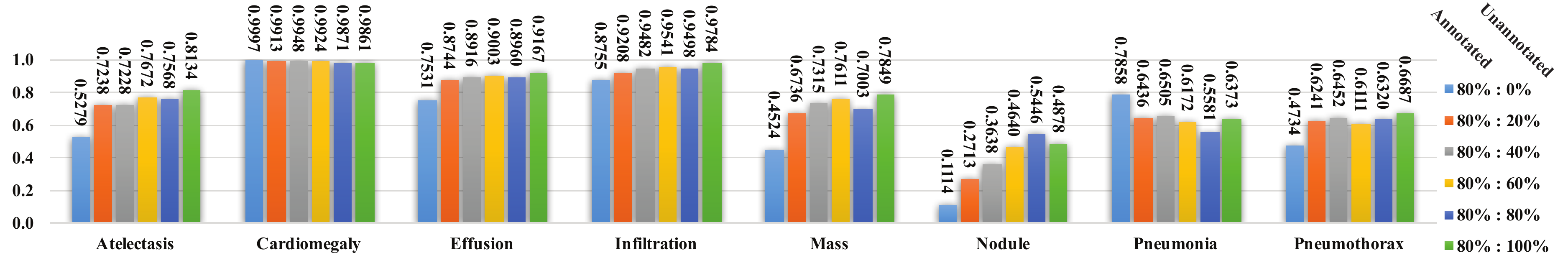}
  \caption{Disease localization accuracy using IoR where T(IoR)=0.1. Training set:  annotated samples, $80\%$ ($704$ images); unannotated samples,  \{$0\%$ ($0$), $20\%$ ($22,248$), $40\%$ ($44,496$), $60\%$ ($66,744$), $80\%$ ($88,892$), $100\%$ ($111,240$)\} from left to right for each disease type. The evaluation set is $20\%$ annotated samples which are not included in the training set. Using annotated samples only can produce a model which localizes some diseases. As the amount of unannotated samples increases in the training set, the localization accuracy is improved and all diseases can be localized. The joint formulation for both types of samples enables unannotated samples to improve the performance with weak supervision.  } 
  \label{fig:localization_ior_2} 
  \vspace{-1.5em}
\end{figure*}

\begin{table*}[t]
    \centering
    \vspace{-1em}
	\resizebox{1.85\columnwidth}{!}{
		\begin{tabular}{c|ccccccccc}
		\hline
T(IoU)               & \multicolumn{1}{c|}{Model} & Atelectasis               & Cardiomegaly              & Effusion                 & Infiltration             & Mass                     & Nodule                  & Pneumonia              & Pneumothorax            \\ \hline
\multirow{2}{*}{0.1} & \multicolumn{1}{c|}{ref.}  & 0.69                      & 0.94                      & 0.66                     & 0.71                     & 0.40                     & 0.14                    & \textbf{0.63}                   & 0.38                    \\ \cline{2-10} 
                     & \multicolumn{1}{c|}{ours}  & $\textbf{0.71}\pm 0.05 $  & $\textbf{0.98} \pm 0.02 $ & $\textbf{0.87} \pm 0.03$ & $\textbf{0.92} \pm 0.05$ & $\textbf{0.71} \pm0.10$  & $\textbf{0.40} \pm0.10$ & $0.60\pm 0.11$        & $\textbf{0.63}\pm0.09$ \\ \hline
\multirow{2}{*}{0.2} & \multicolumn{1}{c|}{ref.}  & 0.47                      & 0.68                      & 0.45                     & 0.48                     & 0.26                     & 0.05                    & 0.35                   & 0.23                    \\ \cline{2-10} 
                     & \multicolumn{1}{c|}{ours}  & $\textbf{0.53}\pm 0.05 $  & $\textbf{0.97} \pm 0.02 $ & $\textbf{0.76} \pm 0.04$ & $\textbf{0.83}\pm 0.06 $ & $\textbf{0.59} \pm 0.10$ & $\textbf{0.29}\pm 0.10$ & $\textbf{0.50}\pm0.12$ & $\textbf{0.51}\pm0.08$  \\ \hline
\multirow{2}{*}{0.3} & \multicolumn{1}{c|}{ref.}  & 0.24                      & 0.46                      & 0.30                     & 0.28                     & 0.15                     & 0.04                    & 0.17                   & 0.13                    \\ \cline{2-10} 
                     & \multicolumn{1}{c|}{ours}  & $\textbf{0.36}\pm 0.08$   & $\textbf{0.94} \pm 0.01$  & $\textbf{0.56}\pm 0.04$  & $\textbf{0.66}\pm 0.07$  & $\textbf{0.45}\pm 0.08$  & $\textbf{0.17}\pm0.10$  & $\textbf{0.39}\pm0.09$ & $\textbf{0.44}\pm0.10$  \\ \hline
\multirow{2}{*}{0.4} & \multicolumn{1}{c|}{ref.}  & 0.09                      & 0.28                      & 0.20                     & 0.12                     & 0.07                     & 0.01                    & 0.08                   & 0.07                    \\ \cline{2-10} 
                     & \multicolumn{1}{c|}{ours}  & $\textbf{0.25}\pm 0.07$   & $\textbf{0.88}\pm 0.06$   & $\textbf{0.37}\pm 0.06$  & $\textbf{0.50}\pm 0.05$  & $\textbf{0.33}\pm 0.08$  & $\textbf{0.11}\pm0.02$  & $\textbf{0.26}\pm0.07$ & $\textbf{0.29}\pm0.06$  \\ \hline
\multirow{2}{*}{0.5} & \multicolumn{1}{c|}{ref.}  & 0.05                      & 0.18                      & 0.11                     & 0.07                     & 0.01                     & 0.01                    & 0.03                   & 0.03                    \\ \cline{2-10} 
                     & \multicolumn{1}{c|}{ours}  & $\textbf{0.14}\pm 0.05$   & $\textbf{0.84}\pm 0.06$   & $\textbf{0.22}\pm 0.06$  & $\textbf{0.30}\pm 0.03$  & $\textbf{0.22}\pm 0.05$  & $\textbf{0.07}\pm0.01$  & $\textbf{0.17}\pm0.03$ & $\textbf{0.19}\pm0.05$  \\ \hline
\multirow{2}{*}{0.6} & \multicolumn{1}{c|}{ref.}  & 0.02                      & 0.08                      & 0.05                     & 0.02                     & 0.00                     & 0.01                    & 0.02                   & 0.03                    \\ \cline{2-10} 
                     & \multicolumn{1}{c|}{ours}  & $\textbf{0.07}\pm 0.03$   & $\textbf{0.73}\pm 0.06$   & $\textbf{0.15}\pm 0.06$  & $\textbf{0.18}\pm 0.03$  & $\textbf{0.16}\pm 0.06$  & $\textbf{0.03}\pm0.03$  & $\textbf{0.10}\pm0.03$ & $\textbf{0.12}\pm0.02$  \\ \hline
\multirow{2}{*}{0.7} & \multicolumn{1}{c|}{ref.}  & 0.01                      & 0.03                      & 0.02                     & 0.00                     & 0.00                     & 0.00                    & 0.01                   & 0.02                    \\ \cline{2-10} 
                     & \multicolumn{1}{c|}{ours}  & $\textbf{0.04}\pm 0.01$   & $\textbf{0.52}\pm 0.05$   & $\textbf{0.07}\pm 0.03$  & $\textbf{0.09}\pm0.02$   & $\textbf{0.11}\pm0.06 $  & $\textbf{0.01}\pm 0.00$ & $\textbf{0.05}\pm0.03$ & $\textbf{0.05}\pm0.03$  \\ \hline
		\end{tabular}
	}
\caption{Disease localization accuracy comparison using IoU where T(IoU)=\{0.1, 0.2, 0.3, 0.4, 0.5, 0.6, 0.7\}.The bold values denote the best results. Note that we round the results to two decimal digits for table readability. Using different thresholds, our model outperforms the reference baseline in most cases and remains capability of localizing diseases when the threshold is big.}
\label{tbl:localization_iou_comparison}
\vspace{-0.5em}
\end{table*}

 \begin{figure*}
    \centering
    \includegraphics[width=1.95\columnwidth]{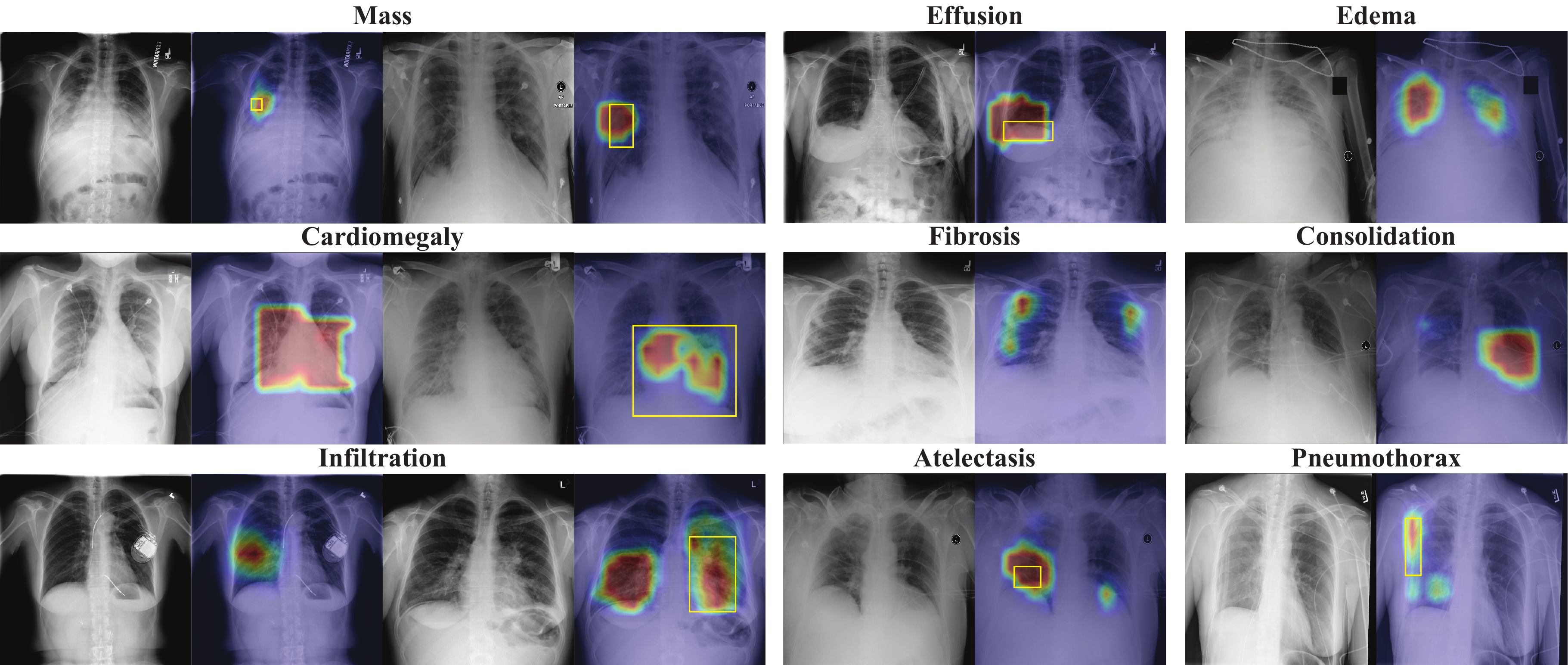} 
    \caption{Example localization visualization on the test images. The visualization is generated by rendering the final output tensor as heatmaps and overlapping on the original images. We list some thoracic diseases as examples. The left image in each pair is the original chest X-ray image and the right one is the localization result. All examples are positive for corresponding labels. We also plot the ground-truth bounding boxes in yellow on the results when they are provided in the dataset.}
    \label{fig:qualitative}
    \vspace{-1.5em}
 \end{figure*}

\subsection{Disease localization}
Similarly, we conduct a 5-fold cross-validation.
For each fold, we have done three experiments. 
In the first experiment, we investigate the importance of bounding box supervision by using all the unannotated images and increasing the amount of annotated images from $0\%$ to $80\%$ by the step of $20\%$ (Figure~\ref{fig:localization_ior_1}).
In the second one, we fix the amount of annotated images to $80\%$ and increase the amount of unannotated images from $0\%$ to $100\%$ by the step of $20\%$ to observe whether unannotated images are able to help annotated images to improve the performance (Figure~\ref{fig:localization_ior_2}).
At last, we train the model with $80\%$ annotated images and half ($50\%$) unannotated images to compare localization accuracy with the reference baseline~\cite{wang2017chestx} (Table~\ref{tbl:localization_iou_comparison}).
For each experiment, the model is always evaluated on the fixed $20\%$ annotated images for this fold.

\textbf{Evaluation metrics.}
We evaluate the detected regions (which can be non-rectangular and discrete) against the annotated ground truth (GT) bounding boxes,
using two types of measurement: intersection over union ratio (IoU) and intersection over the detected region (IoR)
\footnote{Note that we treat discrete detected regions as one prediction region, thus IoR is analogous to intersection over the detected bounding box area ratio (IoBB).}.
The localization results are only calculated for those eight disease types with ground truth provided. 
We define a correct localization when either IoU $>$ T(IoU) or IoR $>$ T(IoR), where T(*) is the threshold.

\textbf{Bounding box supervision is necessary for localization.}
We present the experiments shown in Figure~\ref{fig:localization_ior_1}. The threshold is set as tolerable as T(IoR)=0.1 to show the training data combination effect on the accuracy. Please refer to the supplementary material for localization performance with T(IoU)=0.1, which is similar to Figure~\ref{fig:localization_ior_1}.
Even though the amount of the complete set of unannotated images is dominant compared with the evaluation set ($111,240$ v.s. $176$),
without annotated images (the most left bar in each group), the model fails to generate accurate localization for most disease types. Because in this situation, the model is only supervised by image-level labels and optimized using probabilistic approximation from patch-level predictions. 
As we increase the amount of annotated images gradually from $0\%$ to $80\%$ by the step of $20\%$ (from left to right in each group), the localization accuracy for each type is increased accordingly.
We can see the necessity of bounding box supervision by observing the localization accuracy increase.
Therefore, the bounding box is necessary to provide accurate localization results and the accuracy is positively proportional to the amount of annotated images.
We have similar observations when T(*) varies.

\textbf{More unannotated data does not always mean better results for localization.}
In Figure~\ref{fig:localization_ior_2}, when we fix the amount of annotated images and increase the amount of unannotated ones for training (from left to right in each group), the localization accuracy does not increase accordingly. Some disease types achieve very high accuracy (even highest) without any unannotated images (the most left bar in each group), such as ``Pneumonia'' and ``Cardiomegaly''. 
Similarly as described in the discussion of Section \ref{subsec:classification}, unannotated images and too many negative samples degrade the localization performance for these diseases.
All disease types experience an accuracy increase, a peak score, and then an accuracy fall (from orange to green bar in each group). 
Therefore, with bounding box supervision, unannotated images will help to achieve better results in some cases and it is not necessary to use all of them.

\textbf{Comparison with the reference model.}
In each fold, we use $80\%$ annotated images and $50\%$ unannotated images to train the model and evaluate on the other $20\%$ annotated images in each fold. Since we use 5-fold cross-validation, the complete set of annotated images has been evaluated to make a relatively fair comparison with the reference model.
In Table \ref{tbl:localization_iou_comparison}, we compare our localization accuracy under varying T(IoU) with respect to the reference model in \cite{wang2017chestx}.
Please refer to the supplementary material for the comparison between our localization performance and the reference model with varying T(IoR).
Our model predicts accurate disease regions, not only for the easy tasks like ``Cardiomegaly'' but also for the hard ones like ``Mass'' and ``Nodule'' which have very small regions.
When the threshold increases, our model maintains a large accuracy lead over the reference model. For example, when evaluated by T(IoU)=0.6, our ``Cardiomegaly'' accuracy is still $73.42\%$ while the reference model achieves only $16.03\%$; our ``Mass'' accuracy is $14.92\%$ while the reference model fails to detect any ``Mass'' ($0\%$ accuracy). 
In clinical practice, a specialist expects as accurate localization as possible so that a higher threshold is preferred.
Hence, our model outperforms the reference model with a significant improvement with less training data. 
Please note that as we consider discrete regions as one predicted region, the detected area and its union with GT bboxs are usually larger than the reference work which generates multiple bounding boxes. Thus for some disease types like ``Pneumonia'', when the threshold is small, our result is not as good as the reference.

\subsection{Qualitative results}
Figure \ref{fig:qualitative} shows exemplary localization results of the unified diagnosis model.
The localization enables the explainability of chest X-ray images.
It is intuitive to see that our model produces accurate localization for the diseases compared with the given ground-truth bounding boxes.
Please note for ``Infiltration'' (3th and 4th images in the 3rd row of Figure \ref{fig:qualitative}), both sides of lungs for this patient are infiltrated. Since the dataset only has one bounding box for one disease per image, it misses annotating other bounding boxes for the same disease. Our model gives the remedy.
Even though the extra region decreases the IoR/IoU score in the evaluation, but in clinical practice, it provides the specialist with suspicious candidate regions for further examination.
When the localization results have no ground-truth bounding boxes to compare with, there is also a strong consistency between our results and radiological signs.
For example, our model localizes the enlarged heart region (1st and 2nd images in the 2nd row) which implies ``Cardiomegaly'', and the lung peripheries is highlighted (5th and 6th images in the 2nd row) implying ``Fibrosis'' which is in accordance with the radiological sign of the net-like shadowing of lung peripheries.
The ``Edema'' (7th and 8th images in the 1st row) and ``Consolidation'' (7th and 8th images in the 2nd row) are accurately marked by our model. ``Edema'' always appears in an area that is full of small liquid effusions as the example shows. ``Consolidation'' is usually a region of compressible lung tissue that has filled with the liquid which appears as a big white area. The model successfully distinguishes both diseases which are caused by similar reason.

\section{Conclusion}
We propose a unified model that jointly models disease identification and localization with limited localization annotation data. This is achieved through the same underlying prediction model for both tasks.
Quantitative and qualitative results demonstrate that our method significantly outperforms the state-of-the-art algorithm.



{\small
\bibliographystyle{ieee}
\bibliography{egbib}

\begin{thebibliography}{10}\itemsep=-1pt

\bibitem{tensorflow2015-whitepaper}
M.~Abadi, A.~Agarwal, P.~Barham, E.~Brevdo, Z.~Chen, C.~Citro, G.~S. Corrado,
  A.~Davis, J.~Dean, M.~Devin, S.~Ghemawat, I.~Goodfellow, A.~Harp, G.~Irving,
  M.~Isard, Y.~Jia, R.~Jozefowicz, L.~Kaiser, M.~Kudlur, J.~Levenberg,
  D.~Man\'{e}, R.~Monga, S.~Moore, D.~Murray, C.~Olah, M.~Schuster, J.~Shlens,
  B.~Steiner, I.~Sutskever, K.~Talwar, P.~Tucker, V.~Vanhoucke, V.~Vasudevan,
  F.~Vi\'{e}gas, O.~Vinyals, P.~Warden, M.~Wattenberg, M.~Wicke, Y.~Yu, and
  X.~Zheng.
\newblock {TensorFlow}: Large-scale machine learning on heterogeneous systems,
  2015.
\newblock Software available from tensorflow.org.

\bibitem{akselrod2016region}
A.~Akselrod-Ballin, L.~Karlinsky, S.~Alpert, S.~Hasoul, R.~Ben-Ari, and
  E.~Barkan.
\newblock A region based convolutional network for tumor detection and
  classification in breast mammography.
\newblock In {\em International Workshop on Large-Scale Annotation of
  Biomedical Data and Expert Label Synthesis}, pages 197--205. Springer, 2016.

\bibitem{babenko2008multiple}
B.~Babenko.
\newblock Multiple instance learning: algorithms and applications.

\bibitem{chen2015glaucoma}
X.~Chen, Y.~Xu, D.~W.~K. Wong, T.~Y. Wong, and J.~Liu.
\newblock Glaucoma detection based on deep convolutional neural network.
\newblock In {\em Engineering in Medicine and Biology Society (EMBC), 2015 37th
  Annual International Conference of the IEEE}, pages 715--718. IEEE, 2015.

\bibitem{ieee2008754}
I.~S. Committee et~al.
\newblock 754-2008 ieee standard for floating-point arithmetic.
\newblock {\em IEEE Computer Society Std}, 2008, 2008.

\bibitem{deng2009imagenet}
J.~Deng, W.~Dong, R.~Socher, L.-J. Li, K.~Li, and L.~Fei-Fei.
\newblock Imagenet: A large-scale hierarchical image database.
\newblock In {\em Computer Vision and Pattern Recognition, 2009. CVPR 2009.
  IEEE Conference on}, pages 248--255. IEEE, 2009.

\bibitem{fawcett2006introduction}
T.~Fawcett.
\newblock An introduction to roc analysis.
\newblock {\em Pattern recognition letters}, 27(8):861--874, 2006.

\bibitem{girshick2015fast}
R.~Girshick.
\newblock Fast r-cnn.
\newblock In {\em Proceedings of the IEEE international conference on computer
  vision}, pages 1440--1448, 2015.

\bibitem{girshick2014rich}
R.~Girshick, J.~Donahue, T.~Darrell, and J.~Malik.
\newblock Rich feature hierarchies for accurate object detection and semantic
  segmentation.
\newblock In {\em Proceedings of the IEEE conference on computer vision and
  pattern recognition}, pages 580--587, 2014.

\bibitem{glorot2011deep}
X.~Glorot, A.~Bordes, and Y.~Bengio.
\newblock Deep sparse rectifier neural networks.
\newblock In {\em Proceedings of the Fourteenth International Conference on
  Artificial Intelligence and Statistics}, pages 315--323, 2011.

\bibitem{gylys2017medical}
B.~A. Gylys and M.~E. Wedding.
\newblock {\em Medical terminology systems: a body systems approach}.
\newblock FA Davis, 2017.

\bibitem{he2014spatial}
K.~He, X.~Zhang, S.~Ren, and J.~Sun.
\newblock Spatial pyramid pooling in deep convolutional networks for visual
  recognition.
\newblock In {\em European Conference on Computer Vision}, pages 346--361.
  Springer, 2014.

\bibitem{he2016deep}
K.~He, X.~Zhang, S.~Ren, and J.~Sun.
\newblock Deep residual learning for image recognition.
\newblock In {\em Proceedings of the IEEE conference on computer vision and
  pattern recognition}, pages 770--778, 2016.

\bibitem{he2016identity}
K.~He, X.~Zhang, S.~Ren, and J.~Sun.
\newblock Identity mappings in deep residual networks.
\newblock In {\em European Conference on Computer Vision}, pages 630--645.
  Springer, 2016.

\bibitem{hou2016patch}
L.~Hou, D.~Samaras, T.~M. Kurc, Y.~Gao, J.~E. Davis, and J.~H. Saltz.
\newblock Patch-based convolutional neural network for whole slide tissue image
  classification.
\newblock In {\em Proceedings of the IEEE Conference on Computer Vision and
  Pattern Recognition}, pages 2424--2433, 2016.

\bibitem{huang2017densely}
G.~Huang, Z.~Liu, K.~Q. Weinberger, and L.~van~der Maaten.
\newblock Densely connected convolutional networks.
\newblock In {\em Proceedings of the IEEE conference on computer vision and
  pattern recognition}, volume~1, page~3, 2017.

\bibitem{hwang2016self}
S.~Hwang and H.-E. Kim.
\newblock Self-transfer learning for fully weakly supervised object
  localization.
\newblock {\em arXiv preprint arXiv:1602.01625}, 2016.

\bibitem{ioffe2015batch}
S.~Ioffe and C.~Szegedy.
\newblock Batch normalization: Accelerating deep network training by reducing
  internal covariate shift.
\newblock In {\em International Conference on Machine Learning}, pages
  448--456, 2015.

\bibitem{kingma2014adam}
D.~Kingma and J.~Ba.
\newblock Adam: A method for stochastic optimization.
\newblock {\em arXiv preprint arXiv:1412.6980}, 2014.

\bibitem{liao2017evaluate}
F.~Liao, M.~Liang, Z.~Li, X.~Hu, and S.~Song.
\newblock Evaluate the malignancy of pulmonary nodules using the 3d deep leaky
  noisy-or network.
\newblock {\em arXiv preprint arXiv:1711.08324}, 2017.

\bibitem{liu2017attention}
C.~Liu, J.~Mao, F.~Sha, and A.~L. Yuille.
\newblock Attention correctness in neural image captioning.
\newblock In {\em AAAI}, pages 4176--4182, 2017.

\bibitem{liu2016ssd}
W.~Liu, D.~Anguelov, D.~Erhan, C.~Szegedy, S.~Reed, C.-Y. Fu, and A.~C. Berg.
\newblock Ssd: Single shot multibox detector.
\newblock In {\em European conference on computer vision}, pages 21--37.
  Springer, 2016.

\bibitem{long2015fully}
J.~Long, E.~Shelhamer, and T.~Darrell.
\newblock Fully convolutional networks for semantic segmentation.
\newblock In {\em Proceedings of the IEEE Conference on Computer Vision and
  Pattern Recognition}, pages 3431--3440, 2015.

\bibitem{redmon2016you}
J.~Redmon, S.~Divvala, R.~Girshick, and A.~Farhadi.
\newblock You only look once: Unified, real-time object detection.
\newblock In {\em Proceedings of the IEEE Conference on Computer Vision and
  Pattern Recognition}, pages 779--788, 2016.

\bibitem{ren2015faster}
S.~Ren, K.~He, R.~Girshick, and J.~Sun.
\newblock Faster r-cnn: Towards real-time object detection with region proposal
  networks.
\newblock In {\em Advances in neural information processing systems}, pages
  91--99, 2015.

\bibitem{russakovsky2015imagenet}
O.~Russakovsky, J.~Deng, H.~Su, J.~Krause, S.~Satheesh, S.~Ma, Z.~Huang,
  A.~Karpathy, A.~Khosla, M.~Bernstein, et~al.
\newblock Imagenet large scale visual recognition challenge.
\newblock {\em International Journal of Computer Vision}, 115(3):211--252,
  2015.

\bibitem{shi2017multimodal}
J.~Shi, X.~Zheng, Y.~Li, Q.~Zhang, and S.~Ying.
\newblock Multimodal neuroimaging feature learning with multimodal stacked deep
  polynomial networks for diagnosis of alzheimer's disease.
\newblock {\em IEEE journal of biomedical and health informatics}, 2017.

\bibitem{shin2016learning}
H.-C. Shin, K.~Roberts, L.~Lu, D.~Demner-Fushman, J.~Yao, and R.~M. Summers.
\newblock Learning to read chest x-rays: recurrent neural cascade model for
  automated image annotation.
\newblock In {\em Proceedings of the IEEE Conference on Computer Vision and
  Pattern Recognition}, pages 2497--2506, 2016.

\bibitem{szegedy2015going}
C.~Szegedy, W.~Liu, Y.~Jia, P.~Sermanet, S.~Reed, D.~Anguelov, D.~Erhan,
  V.~Vanhoucke, and A.~Rabinovich.
\newblock Going deeper with convolutions.
\newblock In {\em Proceedings of the IEEE conference on computer vision and
  pattern recognition}, pages 1--9, 2015.

\bibitem{wang2017detecting}
J.~Wang, H.~Ding, F.~Azamian, B.~Zhou, C.~Iribarren, S.~Molloi, and P.~Baldi.
\newblock Detecting cardiovascular disease from mammograms with deep learning.
\newblock {\em IEEE transactions on medical imaging}, 2017.

\bibitem{wang2017chestx}
X.~Wang, Y.~Peng, L.~Lu, Z.~Lu, M.~Bagheri, and R.~M. Summers.
\newblock Chestx-ray8: Hospital-scale chest x-ray database and benchmarks on
  weakly-supervised classification and localization of common thorax diseases.
\newblock In {\em 2017 IEEE Conference on Computer Vision and Pattern
  Recognition (CVPR)}, pages 3462--3471. IEEE, 2017.

\bibitem{wu2015deep}
J.~Wu, Y.~Yu, C.~Huang, and K.~Yu.
\newblock Deep multiple instance learning for image classification and
  auto-annotation.
\newblock In {\em Proceedings of the IEEE Conference on Computer Vision and
  Pattern Recognition}, pages 3460--3469, 2015.

\bibitem{yan2016multi}
Z.~Yan, Y.~Zhan, Z.~Peng, S.~Liao, Y.~Shinagawa, S.~Zhang, D.~N. Metaxas, and
  X.~S. Zhou.
\newblock Multi-instance deep learning: Discover discriminative local anatomies
  for bodypart recognition.
\newblock {\em IEEE transactions on medical imaging}, 35(5):1332--1343, 2016.

\bibitem{zeiler2014visualizing}
M.~D. Zeiler and R.~Fergus.
\newblock Visualizing and understanding convolutional networks.
\newblock In {\em European conference on computer vision}, pages 818--833.
  Springer, 2014.

\bibitem{zhang2017tandemnet}
Z.~Zhang, P.~Chen, M.~Sapkota, and L.~Yang.
\newblock Tandemnet: Distilling knowledge from medical images using diagnostic
  reports as optional semantic references.
\newblock In {\em International Conference on Medical Image Computing and
  Computer-Assisted Intervention}, pages 320--328. Springer, 2017.

\bibitem{zhang2017mdnet}
Z.~Zhang, Y.~Xie, F.~Xing, M.~McGough, and L.~Yang.
\newblock Mdnet: a semantically and visually interpretable medical image
  diagnosis network.
\newblock {\em arXiv preprint arXiv:1707.02485}, 2017.

\bibitem{zhao2016multiscale}
L.~Zhao and K.~Jia.
\newblock Multiscale cnns for brain tumor segmentation and diagnosis.
\newblock {\em Computational and mathematical methods in medicine}, 2016, 2016.

\bibitem{zhou2016learning}
B.~Zhou, A.~Khosla, A.~Lapedriza, A.~Oliva, and A.~Torralba.
\newblock Learning deep features for discriminative localization.
\newblock In {\em Proceedings of the IEEE Conference on Computer Vision and
  Pattern Recognition}, pages 2921--2929, 2016.

\bibitem{zhu2017deep}
W.~Zhu, Q.~Lou, Y.~S. Vang, and X.~Xie.
\newblock Deep multi-instance networks with sparse label assignment for whole
  mammogram classification.
\newblock In {\em International Conference on Medical Image Computing and
  Computer-Assisted Intervention}, pages 603--611. Springer, 2017.

\bibitem{zilly2017glaucoma}
J.~Zilly, J.~M. Buhmann, and D.~Mahapatra.
\newblock Glaucoma detection using entropy sampling and ensemble learning for
  automatic optic cup and disc segmentation.
\newblock {\em Computerized Medical Imaging and Graphics}, 55:28--41, 2017.

\end{thebibliography}
}
\clearpage
\section*{Supplementary material}
\subsection*{Scale patch-level probability to avoid numerical underflow}
Notation $p_{ij}^k$ and $1 - p_{ij}^k$ represent a patch's (the $j$th patch of image $i$) positive and negative probabilities for class $k$.
Their values are always in $[0, 1]$.
We consider the problem of numerical underflow as follows.
The product terms ($\prod$) in Eq.~\ref{eqn:bbox_prod} and Eq.~\ref{eqn:nobbox_prod} can quickly go to $0$ when many of the terms in the product is small due to the limited precision of float numbers. The log loss in Eq.~\ref{eqn:loss} mitigates this for Eq.~\ref{eqn:bbox_prod}, but does not help Eq.~\ref{eqn:nobbox_prod}, since the log function can not directly affect its product term. This effectively renders Eq.~\ref{eqn:nobbox_prod} as a constant value of $1$,
making it irrelevant on updating the network parameters. (The contribution of the  gradient from Eq.~\ref{eqn:nobbox_prod} will be close to $0$.)
Similar things happen at test time. To do binary classification for an image, we determine its label by thresholding the image-level score (Eq.~\ref{eqn:nobbox_prod}).
It is impossible to find a threshold in $[0,1]$ to distinguish the image-level scores when the score (Eq.~\ref{eqn:nobbox_prod}) is a constant of $1$; all the images will be labeled the same.


Fortunately, if we can make sure that the image-level scores $p(y_k|x_i, {\rm bbox}^k_i)$'s and $p(y_k|x_i)$ spread out in $[0, 1]$ instead of congregating at $1$, we then can find an appropriate threshold for the binary classification. 
To this end, we normalize $p_{ij}^k$ and $1 - p_{ij}^k$ from $[0,1]$ to $[0.98, 1]$. The reason of such choice is as follows. In the actual system, we often use single-precision floating-point number to represent real numbers. It can represent a real number as accurate as $7$ decimal digits \cite{ieee2008754}. 
If the number of patches in an image, $m=16 \times 16$, a real number $p \in [0,1]$ should be larger than around $0.94$ (by obtaining $p$ from $p^{256}\geq 10^{-7}$) to make sure that the $p^{m}$ varies smoothly in $[0,1]$ w.r.t. $p$ changes in $[0.94, 1]$.
To be a bit more conservative, we set $0.98$ as our lower limit in our experiment.
This method enables valid and efficient training and testing of our method.
And in the evaluation, the number of thresholds can be finite to calculate the AUC scores, as the image-level probability score is well represented using the values in $[0,1]$. A downside of our approach is that a normalized patch-level probability score does not necessarily reflect the meaning of probability anymore.

\subsection*{Disease Localization Results}
\begin{figure*}
  \centering 
  \includegraphics[width=1.9\columnwidth]{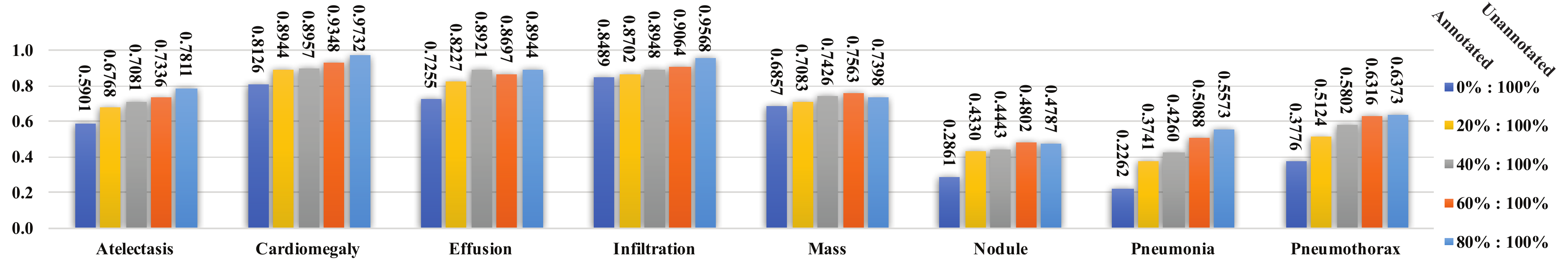}
  \caption{Disease localization accuracy using IoU where T(IoU)=0.1. Training set:  annotated samples, \{$0\%$ ($0$), $20\%$ ($176$), $40\%$ ($352$), $60\%$ ($528$), $80\%$ ($704$)\} from left to right for each disease type; unannotated samples, $100\%$ ($111,240$ images). The evaluation set is $20\%$ annotated samples which are not included in the training set. For each disease, the accuracy is increased from left to right, as we increase the amount of annotated samples, because more annotated samples bring more bounding box supervision to the joint model.} 
  \label{fig:localization_iou_1} 
\end{figure*}

Similarly, we investigate the importance of bounding box supervision by using all the unannotated images and increasing the amount of annotated images from $0\%$ to $80\%$ by the step of $20\%$ (Figure~\ref{fig:localization_iou_1}).
without annotated images (the most left bar in each group), the model is only supervised by image-level labels and optimized using probabilistic approximation from patch-level predictions. The results by unannotated images only are not able to generate accurate localization of disease.
As we increase the amount of annotated images gradually from $0\%$ to $80\%$ by the step of $20\%$ (from left to right in each group), the localization accuracy for each type is increased accordingly.

\begin{figure*}
  \centering 
  \includegraphics[width=1.9\columnwidth]{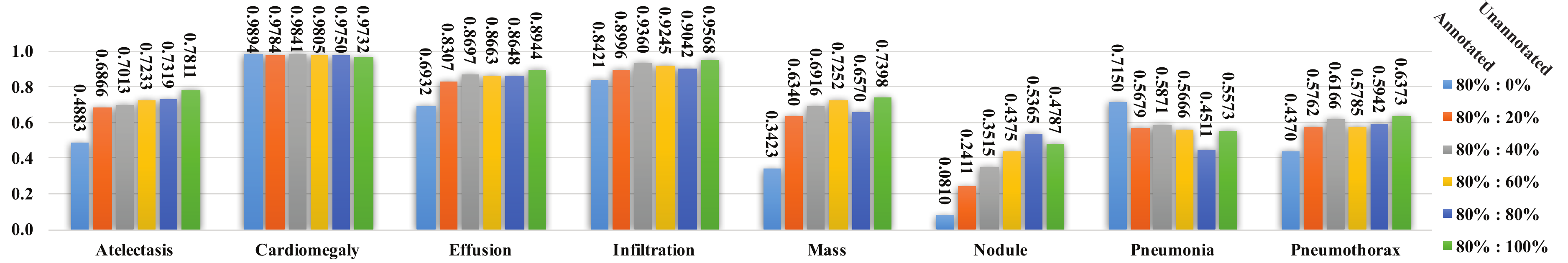}
  \caption{Disease localization accuracy using IoU where T(IoU)=0.1. Training set:  annotated samples, $80\%$ ($704$ images); unannotated samples,  \{$0\%$ ($0$), $20\%$ ($22,248$), $40\%$ ($44,496$), $60\%$ ($66,744$), $80\%$ ($88,892$), $100\%$ ($111,240$)\} from left to right for each disease type. The evaluation set is $20\%$ annotated samples which are not included in the training set. Using annotated samples only can produce a model which localizes some diseases. As the amount of unannotated samples increases in the training set, the localization accuracy is improved and all diseases can be localized. The joint formulation for both types of samples enables unannotated samples to improve the performance with weak supervision. } 
  \label{fig:localization_iou_2} 
\end{figure*}

Next, we fix the amount of annotated images to $80\%$ and increase the amount of unannotated images from $0\%$ to $100\%$ by the step of $20\%$ to observe whether unannotated images are able to help annotated images to improve the performance (Figure~\ref{fig:localization_iou_2}). 
For some diseases, it achieves the best accuracy without any unannotated images.
For most diseases, the accuracy experience an accuracy increase, a peak score, and then an accuracy fall (from orange to green bar in each group) as we increase the amount of unannotated images. A possible explanation is that too many unannotated images overwhelm the strong supervision from the small set of annotated images. A possible remedy is to lower the weight of unannotated images during training.

Lastly, We use $80\%$ annotated images and $50\%$ unannotated images to train the model and evaluate on the other $20\%$ annotated images in each fold. 
Comparing with the reference model \cite{wang2017chestx}, our model achieves higher localization accuracy for various T(IoR) as shown in Table~\ref{tbl:localization_iobb_comparison}.
\begin{table*}
    \centering
	\resizebox{1.8\columnwidth}{!}{
		\begin{tabular}{c|c|cccccccc}
		\hline
T(IoR)                & Model & Atelectasis              & Cardiomegaly             & Effusion                & Infiltration            & Mass                    & Nodule                  & Pneumonia               & Pneumothorax            \\ \hline
\multirow{2}{*}{0.1}  & ref.  & 0.62                     & \textbf{1.00}                    & 0.80                    & 0.91                   & 0.59                    & 0.15                    & \textbf{0.86}           & 0.52                    \\ \cline{2-10} 
                      & ours  & $\textbf{0.77} \pm 0.06 $& $0.99 \pm 0.01$ & $\textbf{0.91}\pm 0.04$ & $\textbf{0.95}\pm 0.05$ & $\textbf{0.75}\pm 0.08$ & $\textbf{0.40}\pm 0.11$ & $0.69\pm 0.09$          & $\textbf{0.68}\pm 0.10$ \\ \hline
\multirow{2}{*}{0.25} & ref.  & 0.39                    & 0.99                     & 0.63                    & 0.80                    & 0.46                    & 0.05                    & \textbf{0.71}                    & 0.34                    \\ \cline{2-10} 
                      & ours  & $\textbf{0.57} \pm 0.09$ & $\textbf{0.99} \pm 0.01$ & $\textbf{0.79}\pm 0.02$ & $\textbf{0.88}\pm 0.06$ & $\textbf{0.57}\pm 0.07$ & $\textbf{0.25}\pm 0.10$ & $0.62\pm 0.05$ & $\textbf{0.61}\pm 0.07$ \\ \hline
\multirow{2}{*}{0.5}  & ref.  & 0.19                     & 0.95                     & 0.42                    & \textbf{0.65}                    & 0.31                    & 0.00                    & 0.48                    & 0.27                    \\ \cline{2-10} 
                      & ours  & $\textbf{0.35}\pm 0.04$  & $\textbf{0.98}\pm 0.02$  & $\textbf{0.52}\pm 0.03$ & $0.62\pm 0.08$ & $\textbf{0.40}\pm 0.06$ & $\textbf{0.11}\pm 0.04$ & $\textbf{0.49}\pm 0.08$ & $\textbf{0.43}\pm 0.10$ \\ \hline
\multirow{2}{*}{0.75} & ref.  & 0.09                     & 0.82                     & 0.23                    & 0.44                    & 0.16                    & 0.00                    & 0.29                    & 0.17                    \\ \cline{2-10} 
                      & ours  & $\textbf{0.20}\pm 0.04$ & $\textbf{0.87}\pm 0.05$   & $\textbf{0.34}\pm 0.06$ & $\textbf{0.46}\pm 0.07$ & $\textbf{0.29}\pm 0.06$ & $\textbf{0.07}\pm 0.04$ & $\textbf{0.43}\pm 0.06$ & $\textbf{0.30}\pm 0.07$ \\ \hline
\multirow{2}{*}{0.9}  & ref.  & 0.07                     & \textbf{0.65}            & 0.14                    & \textbf{0.36}                    & 0.09                    & 0.00                    & 0.23                    & 0.12                    \\ \cline{2-10} 
                      & ours  & $\textbf{0.15}\pm 0.03 $ & $0.59 \pm 0.04$          & $\textbf{0.23}\pm 0.05$ & $0.32\pm 0.07$ & $\textbf{0.22}\pm 0.05$ & $\textbf{0.06}\pm 0.03$ & $\textbf{0.34}\pm 0.04$ & $\textbf{0.22}\pm 0.05$ \\ \hline 
		\end{tabular}
	}
\caption{Disease localization accuracy comparison using IoR where T(IoR)=\{0.1, 0.25, 0.5, 0.75, 0.9\}. The bold values denote the best results. Note that we round the results to two decimal digits for table readability. Using different thresholds, our model outperforms the reference baseline in most cases and remains capability of localizing diseases when the threshold is big. The results for the reference baseline are obtained from the latest update of \cite{wang2017chestx}. }
\label{tbl:localization_iobb_comparison}
\end{table*}

\subsection*{Disease Identification Results using NIH data splits}
We additionally evaluate our method using the updated data split from Wang \etal~\cite{wang2017chestx}.\footnote{This split was posted after CVPR 2018 paper submission deadline.}
Note that in this updated split, all images with annotated bounding boxes are in the test list. Thus, we cannot use any annotation information to improve the training. Also, in this released split, they split the images at the patient level. In Table~\ref{tbl:classification_comparison_2}, we  explore different image models including variants of ResNet~\cite{he2016deep,he2016identity} and DenseNet~\cite{huang2017densely}. There are some AUC performance drop compared with Figure~\ref{fig:classification}. Still, without bounding box information, we can see from Table~\ref{tbl:classification_comparison_2} that our MIL formulation itself helps to improve the AUC performance compared with the baseline. 

\begin{table*}	
    \centering
    \vspace{-1.1em}
	\resizebox{1.7\columnwidth}{!}{
		\begin{tabular}{c|ccccccc}
		\hline\hline
		Disease       & Atelectasis               & Cardiomegaly              & Consolidation             & Edema                     & Effusion                   & Emphysema                 & Fibrosis                   \\ \hline
Wang \etal~\cite{wang2017chestx}                    & 0.7003 & 0.8100 & 0.7032 & 0.8052 & 0.7585 & 0.8330 & 0.7859 \\\hline
ResNet\_v1\_50                  & 0.7261 & 0.8415 & 0.7127 & 0.8109 & 0.7921 & 0.8824 & 0.7695 \\\hline
ResNet\_v2\_50                  & 0.7274 & 0.8357 & 0.7198 & 0.8057 & 0.7886 & 0.8876 & 0.7706 \\\hline
ResNet\_v1\_101                 & 0.7227 & 0.8359 & 0.7166 & 0.7958 & 0.7908 & 0.8663 & 0.7588 \\\hline
ResNet\_v2\_101                 & 0.7182 & 0.8495 & 0.7165 & 0.8109 & 0.7884 & 0.8562 & 0.7616 \\\hline
DenseNet\_121                   & 0.7280 & 0.8477 & 0.7271 & 0.8226 & 0.7824 & 0.7567 & 0.7625 \\\hline
DenseNet\_161                   & 0.7152 & 0.8322 & 0.7189 & 0.8154 & 0.7725 & 0.7435 & 0.7592 \\\hline
DenseNet\_169                   & 0.7291 & 0.8459 & 0.7201 & 0.8035 & 0.7813 & 0.7513 & 0.7605 \\\hline\hline
Disease       & Hernia                    & Infiltration              & Mass                      & Nodule                    & Pleural Thickening         & Pneumonia                 & Pneumothorax               \\ \hline
Wang \etal~\cite{wang2017chestx}                       & 0.8717 & 0.6614 & 0.6933 & 0.6687 & 0.6835 & 0.6580 & 0.7993 \\\hline
ResNet\_v1\_50                  & 0.6933 & 0.6654 & 0.7826 & 0.7014 & 0.7297 & 0.6519 & 0.8101 \\\hline
ResNet\_v2\_50                  & 0.6933 & 0.6722 & 0.7758 & 0.6961 & 0.7373 & 0.6494 & 0.8075 \\\hline
ResNet\_v1\_101                 & 0.6852 & 0.6612 & 0.7759 & 0.7063 & 0.7380 & 0.6239 & 0.8074 \\\hline
ResNet\_v2\_101                 & 0.6910 & 0.6715 & 0.7830 & 0.7092 & 0.7361 & 0.6532 & 0.8069 \\\hline
DenseNet\_121                   & 0.6532 & 0.6452 & 0.7468 & 0.7017 & 0.7354 & 0.6322 & 0.8015 \\\hline
DenseNet\_161                   & 0.6453 & 0.6549 & 0.7582 & 0.7005 & 0.7309 & 0.6258 & 0.7941 \\\hline
DenseNet\_169                   & 0.6679 & 0.6725 & 0.7426 & 0.7104 & 0.7298 & 0.6325 & 0.7925 \\\hline
		\end{tabular}
	}
\caption{AUC scores comparison among different image models and the baseline. Results are rounded to four decimal digits for accurate comparison. The data split for training and test set is obtained from the latest update of \cite{wang2017chestx}.}
\label{tbl:classification_comparison_2}
\end{table*}

\end{document}